\documentclass[letterpaper, 10 pt, conference]{ieeeconf}  

\IEEEoverridecommandlockouts                              
\usepackage{amssymb}
\usepackage{amsmath}
\usepackage{graphicx}
\graphicspath{{images/}}
\usepackage{float}
\usepackage{hyperref}

\newtheorem{definition}{Definition}

\usepackage{lipsum}
\usepackage{booktabs}

\usepackage[backend=biber]{biblatex}
\newcommand{\R}{\mathbb{R}}
\newcommand{\vv}{\mathbf{v}}

\newcommand{\p}{\mathbf{p}}
\newcommand{\w}{\boldsymbol{\omega}}
\newcommand{\q}{\mathbf{q}}
\newcommand{\x}{\mathbf{x}}
\newcommand{\uu}{\mathbf{u}}
\newcommand{\n}{\mathbf{n}}
\newcommand{\dq}{\dot{\mathbf{q}}}
\newcommand{\ddq}{\ddot{\mathbf{q}}}
\newcommand{\tauu}{\boldsymbol{\tau}}
\newcommand{\spin}{\eta_{\text{spin}}}
\newcommand{\alphan}{\alpha_{\text{nose}}}
\newcommand{\fext}{\mathbf{F}_{\text{ext}}}
\newcommand{\tauext}{\boldsymbol{\tau}_{\text{ext}}}
\newcommand{\xx}{\mathbf{x}}

\title{\LARGE \bf
Throwing a Tight Spiral American Football by a Humanoid Robot
}

\author{Zaid Mahboob$^{1}$ and Bowen Weng$^{1}$
\thanks{$^{1}$Zaid Mahboob and Bowen Weng are with the Department of Computer Science, Iowa State University
        \texttt{\{zaidm,bweng\}@iastate.edu} (corresponding author: Zaid Mahboob)}
\thanks{This work was supported in part by the Presidential Interdisciplinary Research Seed Grant at Iowa State University and in part by the National Science Foundation under Grant FRR-2616778.}
\thanks{Accepted to the 25th IEEE‑RAS International Conference on Humanoid Robots (Humanoids 2026).}
}

\begin{document}

\maketitle
\thispagestyle{empty}
\pagestyle{empty}

\begin{abstract}
Accurate throwing of the American football requires precise regulation of release conditions, where coupled linear and angular momentum determine flight stability and targeting accuracy. While prior work on robotic object throwing has largely focused on generating dynamically feasible release velocities using open-gripper paradigms, explicit control of spin injection at detachment remains underexplored, particularly for aerodynamically anisotropic objects like the American football. In this paper, we present the spin-stabilized controlled tight spiral throw of an American football by a humanoid robot. Achieving this requires (i) accurately reaching the desired coupled momentum, which often involves high degrees-of-freedom (DoF) movements completed within approximately half a second, and (ii) managing the complex transient contact dynamics that arise during the sub-100-millisecond release phase, when the football is effectively underactuated as it moves partially across the fingers. To this end, we develop a coupled whole-body control strategy where the lower body is performing informed stabilization while the upper body is further divided into two phases with (i) a \emph{throw phase} accelerating the football to a target state through trajectory optimization and tracking, and (ii) a \emph{follow-through} phase utilizing model predictive control to actively control the wrist and remaining in-contact fingers. The proposed framework is empirically validated on a 29-DoF Unitree G1 humanoid equipped with a 7-DoF Dex3-1 three-fingered gripper. The thrown American football reaches up to 93.6\% spin efficiency and a 0.286 radians linear-velocity-to-nose-alignment (nose-angle) error (where an ``ideal'' tight spiral corresponds to 100\% spin efficiency and 0 radians nose-angle error) at up to a 5.35~m/s linear velocity and an angular velocity of 14.5~rad/s. A video demonstration is available at \url{https://youtu.be/gw1AeG-Y614}.
\end{abstract}

\section{Introduction}\label{sec:intro}

Among many agile and dynamic bio-inspired robotic tasks, such as dancing~\cite{zhang2026dancing}, martial arts~\cite{xie2025kungfubot}, goalkeeping~\cite{huang2023goalkeeping}, parkour~\cite{wu2026parkour, rudin2026parkour}, table tennis~\cite{durr2026tt}, baseball~\cite{rai_baseball}, basketball~\cite{toyota_basketball}, tennis~\cite{zhang2026tennis}, badminton~\cite{ma2025badminton}, skateboard riding~\cite{han2026husky}, and soccer~\cite{riedmiller2009soccer,robocup}, object throwing occupies a distinct role. At first glance, throwing appears simple: an object is accelerated and then released with a target linear and angular velocity. However, in the pursuit of accuracy, efficiency, and stability, such as in athletic contexts, throwing often involves a coordinated synthesis of linear and angular momentum~\cite{toyota_basketball, rai_baseball, liumomentum2025,spinstability2022}. Among many robotic athletic throwing tasks (e.g., basketball~\cite{toyota_basketball}, baseball~\cite{rai_baseball}) and similar activities like batting~\cite{jia2019batting}, throwing an American football presents a unique challenge. Its elongated, prolate-spheroidal shape induces tightly coupled translational and rotational dynamics, making the release velocity, orientation, and spin critical for flight stability, energy efficiency, and targeting accuracy~\cite{price2020mit, spinstability2022, rae2002}. 

Compounding this difficulty, to the best of our knowledge, there is no established scientific consensus or rigorous quantitative characterization of what constitutes a ``tight spiral'', which physical variables govern its formation, or why particular release strategies produce it. This limited understanding is reflected, in part, in the substantial variation in throwing mechanics among professional quarterbacks (i.e., the players primarily responsible for throwing the football), with no two athletes employing precisely the same technique~\cite{ivy1995apparatus}. This reliance on tacit human expertise presents a significant challenge for robotics: imitation-learning approaches that have proven effective for many manipulation tasks~\cite{zhao2023imitation, chi2025diffusion, shafiullah2022behavior} may be fundamentally limited when neither a unique reference motion nor a well-defined characterization of successful performance is available.

Motivated by these challenges, this paper makes, to the best of our knowledge, the first attempt in robotics to investigate how to enable a humanoid robot to throw an American football with a stable, tight spiral.

\begin{figure}[t]
    \vspace{2mm}
    \centering
    \includegraphics[width=\columnwidth]{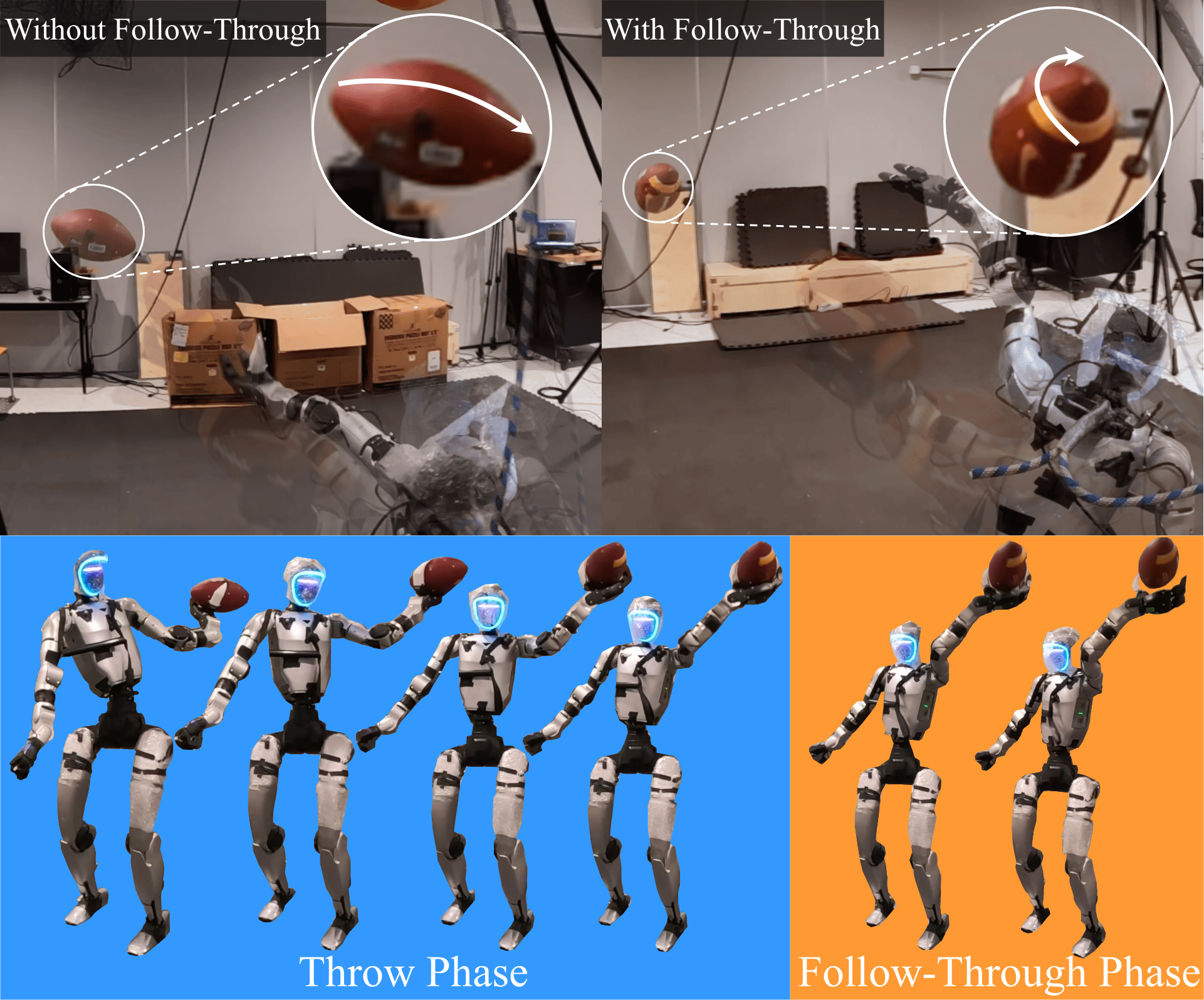}
    \vspace{-7mm}
    \caption{
    Demos of a humanoid robot (Unitree G1 with Dex3-1) throwing a tight spiral American football. The controller has two sequenced phases: the throw phase generates the coupled momentum, while the follow-through phase uses model predictive control (MPC) to regulate the release dynamics and stabilize the tight spiral. Without follow-through (top left), the ball tumbles due to an unstable release; with follow-through, the ball has a stable tight spiral (top right).
    }
    \vspace{-12mm}
    \label{fig:introduction}
\end{figure}

\subsection{Literature Review}
\label{subsec:lit-review}
In the world of robotic throwing, the primary focus has been on generating dynamically feasible release conditions using jaw grippers, where the object is accelerated, followed by a release (instantaneous gripper opening)~\cite{liu2022solution}. Some works generate dynamically feasible throwing motions through kinodynamic planning and trajectory optimization~\cite{kim2010motion,sintov2015stochastic,lombai2009throwing}. Following a similar release mechanism, recent studies have also considered learning-based and data-driven approaches~\cite{zeng2020tossingbot,aslam2025dartbot, kasaei2023throwing}. Despite substantial progress in robotic throwing, most existing methods do not seek precise control over coupled linear and angular momentum~\cite{zeng2020tossingbot, aslam2025dartbot}. They also neglect the transient release dynamics unfolding during the final milliseconds of contact~\cite{aslam2025dartbot,zeng2020tossingbot}. A recent study by Liu and Billard~\cite{liu2025sliding} investigated these release dynamics for a planar two-finger jaw gripper. However, such models are limited to a planar, two-dimensional setting with two symmetric contact points and fail to capture the complex, three-dimensional release dynamics of multi-fingered manipulation tasks, such as throwing an American football, where the contacts undergo coupled rolling and sliding.

Another group of studies related to this paper's proposal focuses on batting tasks, such as robotic table tennis~\cite{durr2026tt, xiong2012tt, dambrosio2025googlett}, badminton~\cite{ma2025badminton}, and tennis~\cite{zaidi2023tennis, zhang2026tennis}, where linear and angular momentum are transferred through a brief impulse. Prior robotic batting systems predict the incoming ball and execute predefined strokes~\cite{zaidi2023tennis}, or employ hierarchical sim-to-real reinforcement learning~\cite{dambrosio2025googlett,zhang2026tennis,durr2026tt}, often bypassing explicit ball-racket impact modeling. When considered, contact is typically estimated as a single instantaneous collision that neglects friction, spin transfer, deformation, and finite contact duration~\cite{ma2025badminton}. In contrast, throwing an American football imparts coupled linear and angular momentum through three-dimensional, multi-finger contact, making the underlying multi-contact dynamics essential to successful execution.

Among the biomechanical tasks discussed above, consistently throwing an American football tight spiral is challenging even for skilled humans. It demands coordinating a 50-ms arm-acceleration phase with a brief release window~\cite{laliberty2012sportsball,rafel2009throwingbiomechanics}. Furthermore, to the best of our knowledge, there is no scientific understanding of underlying mechanics, as reflected in the highly individualized techniques of elite quarterbacks~\cite{ivy1995apparatus}. This reliance on tacit expertise severely limits imitation learning~\cite{zhao2023imitation, chi2025diffusion, shafiullah2022behavior}, which requires unique reference trajectories and clear mathematical success criteria. Broader machine-learning approaches face additional challenges because the transient release involves complex rolling and sliding across multiple contacts. These interactions are difficult to model accurately, which can produce a substantial sim-to-real gap.

\subsection{Main Contributions}\label{sec:contribution}
To the best of our knowledge, this paper makes the first attempt of enabling a spin-stabilized tight spiral throw of the American football through a humanoid robot. The main contributions of this paper are summarized as follows:
\begin{itemize}
    \item \textbf{Modeling}: We formally define the tight spiral through quantifiable metrics and present a contact-aware dynamics model of the football.
    \item \textbf{Control}: We propose a whole-body controller that combines informed lower-body stabilization with a two-stage upper-body strategy: a throw phase that generates coupled momentum, followed by a model predictive control (MPC)-based follow-through phase to regulate transient spin-stabilized release and prevent contact-induced disturbances.
    \item \textbf{Real-World Experiment}: We implement the proposed framework on a Unitree G1 humanoid equipped with a three-finger Dex3-1 gripper augmented by a custom thumb ``nail'' extension for clean detachment in real-world experiments. We demonstrate a spin-stabilized tight spiral throw and compare it against a non-stabilized baseline that does not explicitly control the transient finger–ball contact dynamics during release.
\end{itemize}

\section{Preliminaries and Problem Formulation}\label{sec:prob}

\noindent\textbf{Notation.} The set of real numbers is denoted by $\mathbb{R}$, and the special orthogonal group in 3D by $\text{SO}(3)$. The standard $\ell_2$ norm is denoted by $\|\cdot\|$, scalar absolute value by $|\cdot|$, and the skew-symmetric cross-product matrix of a vector by $[\cdot]_{\times}$.

\subsection{American Football Model and Tight Spiral}
\label{subsec:football}
\begin{figure}[b]
    \vspace{-10mm}
    \centering
    \includegraphics[width=0.9\columnwidth]{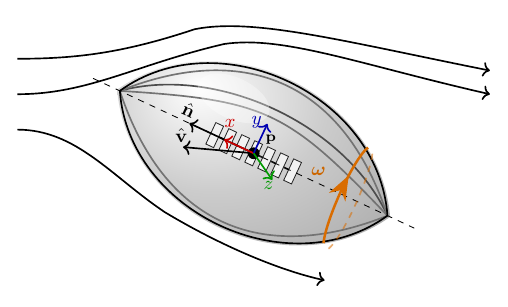}
    \vspace{-7mm}
    \caption{Football rigid-body state and kinematic quantities used in the tight spiral Definition~\ref{def:tight_spiral}. The football state is defined as $\xx_{\text{b}} = (\p, R, \vv, \w)$, where $\p, \vv, \w \in \R^{3}$ represent the center-of-mass position, linear velocity, and angular velocity, respectively, and $R \in \text{SO}(3)$ is the football orientation matrix mapping vectors from the body frame to the world frame. The football longitudinal body-frame axis is defined as $\mathbf{e}_x = [1, 0, 0]^\top$, and the corresponding world-frame nose direction is given by $\hat{\n} = R\mathbf{e}_x$.}
    \label{fig:football_kinematics}
\end{figure}

The \emph{American football} (referred to as \emph{football} for simplicity in the remainder of this paper) is modeled as a rigid body, as shown in Fig.~\ref{fig:football_kinematics}. Rotational symmetry is assumed about the longitudinal axis, a reasonable approximation for a typical football because its mass is distributed nearly uniformly around this axis.

The football's translational and rotational dynamics are modeled using the Newton-Euler equations: 
\begin{equation}
\label{eq:football_dynamics}
\begin{alignedat}{2}
\dot{\p} &= \vv, &\hspace{2mm} \dot{\vv} &= \fext/m + \mathbf{g},\\
\dot{R} &= [\w]_\times R, & \hspace{2mm} \dot{\w} &= (RIR^\top)^{-1} ( \tauext - \w{\times}(RIR^\top\w) ).
\end{alignedat}
\end{equation}

In the football dynamics~\eqref{eq:football_dynamics}, $m$ is the mass of the football, $I$ is the inertia matrix, and $\mathbf{g}=[0,0,-9.81]^\top~\text{m}/\text{s}^{2}$ denotes the gravitational acceleration. The vectors $\fext,\,\tauext\in\R^{3}$ denote the resultant external force and torque acting on the football, respectively. Note that such resultant external forces and torques can be generated through a variety of mechanisms, such as a football launcher~\cite{hollaus2021launcher} and the multi-contact interaction with an articulated robotic hand. A ``tight spiral'' throw~\cite{price2020mit} is thus formally defined as follows:
\begin{definition}[\textbf{Tight Spiral}]
\label{def:tight_spiral}
A football state $\xx_{\text{b}}$ is a \emph{tight spiral} if and only if
\begin{equation}
\label{eq:tight_spiral_conditions}
\|\vv\|\!>\!0, \, \|\w\|\!>\!0, \, \hat{\n}\!\times\!\vv\!=\!\mathbf{0}, \, \hat{\n}^{\top}\!\vv\!>\!0, \, \hat{\n}\!\times\!\w\!=\!\mathbf{0}.
\end{equation}
\end{definition}

The above conditions~\eqref{eq:tight_spiral_conditions} for tight spiral require non-zero velocities aligned with the football’s longitudinal axis. A tight spiral is important for football throwing as it produces a stable, nose-forward~\cite{price2020mit}, and an easily predictable trajectory~\cite {dolgov2009evidence}. Note the above condition concerns only the directions of the vectors, not their magnitudes, i.e., neither the football's linear speed nor its angular speed affects the tight-spiral quality as characterized by Definition~\ref{def:tight_spiral}. In practice, however, these magnitudes do influence the football's flight dynamics through aerodynamic effects~\cite{price2020mit}, which are beyond the scope of this paper. Definition~\ref{def:tight_spiral} is further abstracted into two concrete metrics derived from practically observable states as
\begin{equation}
\label{eq:spiral_metrics}
\spin(\hat{\n},\w)\!=\!\frac{|\hat{\n}^\top \w|}{\|\w\|},
\alphan(\hat{\n},\vv)\!=\!\arccos \left( \frac{|\hat{\n}^\top \vv|}{\|\vv\|} \right),
\end{equation}

where $\spin$ measures the alignment between the nose axis and angular velocity, while $\alphan$ measures the alignment between the nose axis and linear velocity. For $\spin=1$ and $\alphan=0$, the corresponding states concur the ``ideal'' tight spiral as per Definition~\ref{def:tight_spiral}. In this paper, among many possible ways to enable a tight spiral throw (e.g., using a football launcher~\cite{hollaus2021launcher}), we focus on adopting a humanoid robot with a multi-finger gripper. 

\subsection{Humanoid Robot Model}
\label{subsec:humanoid}

Formally speaking, the humanoid robot is modeled as a floating-base rigid-body system that generates the throwing motion while maintaining support contacts with the ground as
\begin{equation}
\label{eq:humanoid_dynamics}
\begin{aligned}
    &M(\q)\ddot{\q} + C(\q,\dq)\dq + G(\q) \\
    &\qquad = \underbrace{S^{\top}\tauu}_{\text{joint actuation}} \!\!+\! \underbrace{\sum_{j=1}^{2} J_{c_j}^{\top}(\q)\mathbf{f}^{j}}_{\text{foot-ground contacts}} \!+\!\! \underbrace{\sum_{i=1}^{n_h} J_{h_i}^{\top}(\q)\mathbf{f}_{h}^{i}}_{\text{fingertip-football contacts}}.
\end{aligned}
\end{equation}
Its generalized coordinates and velocities are $\q,\dq\in\R^{6+n}$, where the first six coordinates represent the floating base and the remaining $n$ coordinates represent the actuated joints. The associated actuated torques are denoted by $\tauu \in \R^{n}$. Here, $M(\q)\in\R^{(6+n)\times(6+n)}$ denotes the generalized mass-inertia matrix, $C(\q,\dq)\dq\in\R^{6+n}$ the Coriolis and centrifugal terms, $G(\q)\in\R^{6+n}$ the gravity vector, and $S\in\R^{n\times(6+n)}$ the actuator selection matrix. The two foot-ground contacts are represented by $J_{c_j}(\q)\in\R^{6\times(6+n)}$ and $\mathbf{f}^{j}\in\R^{6}$, denoting the spatial contact Jacobian and contact wrench of the $j$-th foot, respectively. During the throwing motion, the football is held through multiple fingertip contacts. For the $i$-th fingertip contact, $J_{h_i}(\q)\in\R^{3\times(6+n)}$ denotes the translational Jacobian, and $\mathbf{f}_{h}^{i}\in\R^{3}$ denotes the contact force exerted by the football on the fingertip, where $n_h$ denotes the number of active fingertip contact points.

\section{Main Method}\label{sec:method}
Among many control paradigms available for humanoid robots in enabling a tight spiral throw, we adopt a coupled architecture comprising two interacting control modules: (i) a legged controller $\pi_{\text{leg}}$ that maintains stable ground support, and (ii) a manipulation controller $\pi_{\text{mnp}}$ that executes the throwing motion. The two controllers exchange state information. The legged controller $\pi_{\text{leg}}$ receives the manipulation state, allowing it to anticipate disturbances induced by the throwing motion. The manipulation controller $\pi_{\text{mnp}}$ receives an estimate of the pelvis motion from an extended Kalman filter (EKF)~\cite{hartley2020ekf} based on contact, leg kinematics, and IMU measurements. This modular design supports high-frequency computational efficiency while allowing different modules to execute at frequencies suited to their respective needs.

\begin{figure*}[t]
    \centering
    \vspace{2mm}
    \includegraphics[width=0.85\textwidth]{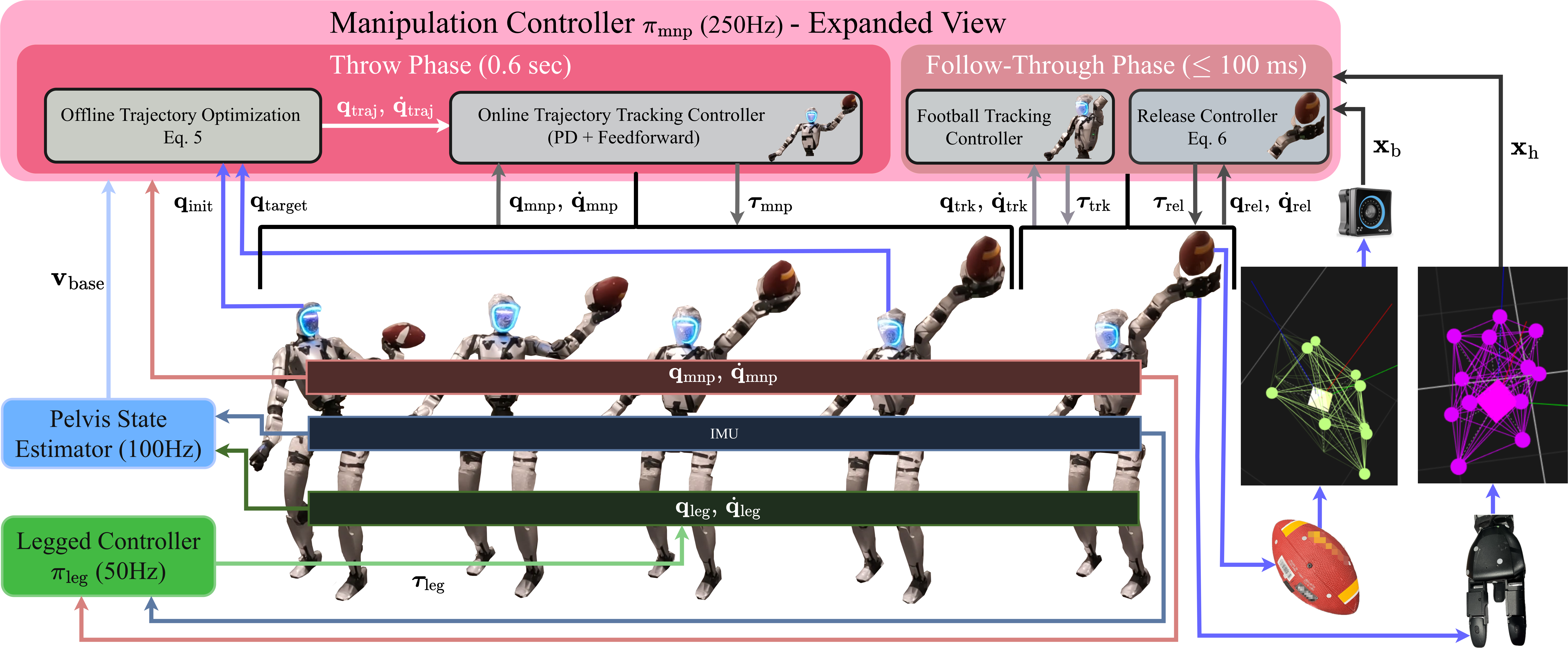}
    \vspace{-4mm}
    \caption{The proposed whole-body control architecture along with the estimation solutions for the humanoid football throwing of a tight spiral throw.
    }
    \vspace{-7mm}
    \label{fig:whole_body_control_architecture}
\end{figure*}

In this work, the legged controller $\pi_{\text{leg}}$ is flexible and may adopt a variety of model-based~\cite{carlo2018cheetah, kajita2003zmp} and learning-based designs~\cite{kumar2021rma, hwangbo2019anymal, li2025cassie, bjorck2025gr00t}. Although the performance of these designs varies, the resulting pelvis state $\mathbf{v}_{\text{base}}$ is estimated using an EKF and provided to the manipulation controller $\pi_{\text{mnp}}$. For the manipulation controller $\pi_{\text{mnp}}$, we adopt a two-phased control strategy shown in Fig.~\ref{fig:whole_body_control_architecture}: (i) a throw phase that accelerates the football and (ii) a follow-through phase that regulates the transient release. This decomposition is in part inspired by human throwing, in which the arm first accelerates the football and the fingers then coordinate its release~\cite{smith2017electromyographic}. During this interval, residual finger-football contact can alter the ball's release state. If not properly regulated, this residual contact can induce the wobble as illustrated in the top-left of Fig.~\ref{fig:introduction}.

\subsection{Throw Phase}
\label{sec:throw_phase}

Given a fixed ready-to-throw configuration $\x_0 = [\q_{\text{init}}^{\top},\mathbf{0}^{\top}]^{\top}$ in which the football is securely grasped, we generate the throwing trajectory using offline trajectory optimization. We enforce that the gripper maintains a non-slip grasp on the football and therefore assume that the football remains rigidly attached to the hand throughout the throw phase. Under this assumption, the trajectory optimization problem is formulated as
\begin{equation}
\label{eq:trajopt}
\begin{aligned}
\min_{\x(\cdot), \uu(\cdot)}  & \int_{0}^{T} \!\!
\mathbf{w}_{\text{run}}
\begin{bmatrix} J_{\text{ref}}(t) & J_{\tau}(t) & J_{\text{acc}}(t) & J_{\text{lim}}(t) \end{bmatrix}^{\top} dt \\
& + \int_{t_a}^{T} \text{w}_{\text{align}} J_{\text{align}}(t) dt \\
& + \mathbf{w}_{\text{term}}
\begin{bmatrix} J_{\text{vel}}(T) & J_{\text{ori}}(T) & J_{\text{pose}}(T) \end{bmatrix}^{\top} \\
\text{s.t.} \quad & \dot{\x}(t) = f(\x(t),\uu(t)), \quad \forall t \in [0,T], \\
& \lvert\uu(t)\rvert \leq \uu_{\max}, \quad \x(0) = [\q_{\text{init}}^{\top},\mathbf{0}^{\top}]^{\top}.
\end{aligned}
\end{equation}

In the trajectory optimization~\eqref{eq:trajopt}, the state and control inputs at time $t$ are defined as $\x(t)=[\q(t)^{\top},\dq(t)^{\top}]^{\top}$ and $\uu(t)=\tauu(t)$, respectively. The formulation seeks to optimize a composite objective by penalizing trajectory deviations from a reference throwing trajectory $\x_{\text{ref}}$ spanning from the initial pose to the release pose ($J_{\text{ref}}(t) = \left\| \x(t) - \x_{\text{ref}}(t) \right\|^2$). It also regularizes the applied joint torques ($J_{\tau}(t) = \left\| \uu(t) \right\|^2$) and accelerations ($J_{\text{acc}}(t) = \left\| \ddq(t) \right\|^2$) while strongly discouraging proximity to joint position and velocity limits ($J_{\text{lim}}(t)$). Over the terminal alignment window $t \in [t_a, T]$, where $T$ is the total time horizon, the objective minimizes misalignment between the nose vector and the linear velocity vector ($J_{\text{align}}(t) = \left\| \hat{\n}(t) \times \hat{\vv}(t) \right\|^2$). At time $T$, the formulation optimizes release conditions by penalizing errors in the target linear and angular velocities ($J_{\text{vel}}(T) = \left\| \vv(T) - \vv^* \right\|^2 + \left\| \boldsymbol{\omega}(T) - \boldsymbol{\omega}^* \right\|^2$), the target release orientation ($J_{\text{ori}}(T) = \left\| \log(R_{\text{target}}^{\top}R(T)) \right\|^2$), and the target joint pose ($J_{\text{pose}}(T) = \left\| \q(T) - \q_{\text{target}} \right\|^2$). Here, $\vv^*$, $\boldsymbol{\omega}^*$, $R_{\text{target}} \in \text{SO}(3)$, and $\q_{\text{target}}$ denote the target terminal linear velocity, angular velocity, orientation, and joint position, respectively. The optimization constraints strictly enforce the rigid-body dynamics, torque limits, and the initial ready-to-throw state.

The optimized trajectory is then tracked online using a proportional–derivative controller with feedforward torque compensation. To compensate for pelvis motion, the manipulation controller subtracts the EKF-estimated pelvis-induced velocity from the commanded motion. At the end of the throw phase, at $t = T$, the thumb opens to initiate release, while the remaining active fingers continue to interact with the football. The subsequent follow-through phase regulates these transient contacts till complete detachment.

\subsection{Follow-Through Phase}
\label{sec:follow_through_mpc}
Without a controlled release, such as simultaneously opening all fingers at the end of the throw, the spiral quality degrades substantially before the football fully detaches from the hand (see top-left of Fig.~\ref{fig:introduction} and Fig.~\ref{fig:spiral_comparison}), making a tight spiral difficult to achieve. To properly control the release during follow-through, the manipulation controller $\pi_{\text{mnp}}$ is divided into a football-tracking controller and an MPC-based release controller that execute in parallel, as shown in Fig.~\ref{fig:whole_body_control_architecture}. The tracking controller commands the arm to maintain multi-finger contact by following the football's trajectory, while the release controller simultaneously ensures a clean detachment to stabilize the tight spiral. Accordingly, we partition the controlled joints as $\q=[\q_{\text{trk}}^{\top},\q_{\text{rel}}^{\top}]^{\top}$, where the football-tracking controller regulates the waist, shoulder, and elbow joints in $\q_{\text{trk}}$, whereas the release controller regulates the wrist and active finger joints in $\q_{\text{rel}}$.

At the current time step $t$, let $\xx_{\text{b}}(t)$, $\xx_{\text{h}}(t)$ and $\q_{\text{rel}}(t)$ denote the measured states of the football, hand, and the release joints, respectively. Over a look-ahead horizon of $H$ steps, we optimize the wrist-finger joint-velocity control sequence $U = (\uu_{\text{rel}, 0}, \ldots, \uu_{\text{rel}, H-1})$, where $\uu_{\text{rel}, k} = \dq_{\text{rel}, k}$. The release controller's MPC is formulated as
\begin{equation}
\label{eq:follow_through_mpc}
\begin{aligned}
\min_U \quad & \sum_{k=0}^{H-1} \mathbf{w}_{\text{mpc}}
\begin{bmatrix}
J_{\text{w},k} & J_{\text{v},k} & J_{\text{u},k} & J_{\text{c},k}
\end{bmatrix}^{\top} \\
\text{s.t.} \quad & \q_{\text{rel},k+1} = \q_{\text{rel},k} + \uu_{\text{rel},k}\Delta t, \\
& \xx_{\text{b},k+1} = f_{\text{b}}\left(\xx_{\text{rel},k}, \uu_{\text{rel},k}, \xx_{\text{r},k}\right), \\
& \lvert\q_{\text{rel},k}\rvert \leq \q_{\text{rel},\max}, \quad \lvert\uu_{\text{rel},k}\rvert \leq \dq_{\text{rel},\max}, \\
& \lvert\tauu_{\text{rel},k}\rvert \leq \tauu_{\text{rel},\max}, \quad \xx_{\text{b},0} = \xx_{\text{b}}(t), \\ 
& \q_{\text{rel},0} = \q_{\text{rel}}(t), \quad \forall k \in \{0, \ldots, H-1\}.
\end{aligned}
\end{equation}

Here, for each prediction step $k$, $\xx_{\text{rel},k} = (\xx_{\text{b},k}, \q_{\text{rel},k})$ represents the combined system state. The wrist-root state commanded by the tracking controller is denoted by $\xx_{\text{r},k}$, with $\xx_{\text{r},0}$ computed from the measured hand state $\xx_{\text{h}}(t)$ and forward kinematics. The formulation penalizes off-axis angular velocity to suppress wobble ($J_{\text{w},k} = \|\w_{\perp,k}\|^2 / (\omega_{\text{ax}}^2 + 1)$), where $\omega_{\text{ax}}$ is the initial signed axial spin measured at the start of the horizon ($k=0$). It also minimizes misalignment between the football nose vector $\hat{\n}_{k}$ and its linear velocity vector $\hat{\vv}_{k}$ ($J_{\text{v},k} = 1 - (\hat{\n}_{k}^{\top}\hat{\vv}_k)^2$). The relative wrist-finger joint velocities are regularized to promote smooth control commands ($J_{\text{u},k}$). Inward fingertip velocity components $v_{i,k}$ that exceed the safe threshold $v_{\text{safe}}$ are penalized ($J_{\text{c},k} = \sum_i \max(0, v_{i,k} - v_{\text{safe}})^2$), thereby mitigating impact forces that could destabilize the release. Each optimization horizon is initialized with the measured football and joint states, $\xx_{\text{b}}(t)$ and $\q_{\text{rel}}(t)$, respectively.

\begin{figure}[t]
    \centering
    \vspace{2mm}
    \includegraphics[width=0.9\columnwidth]{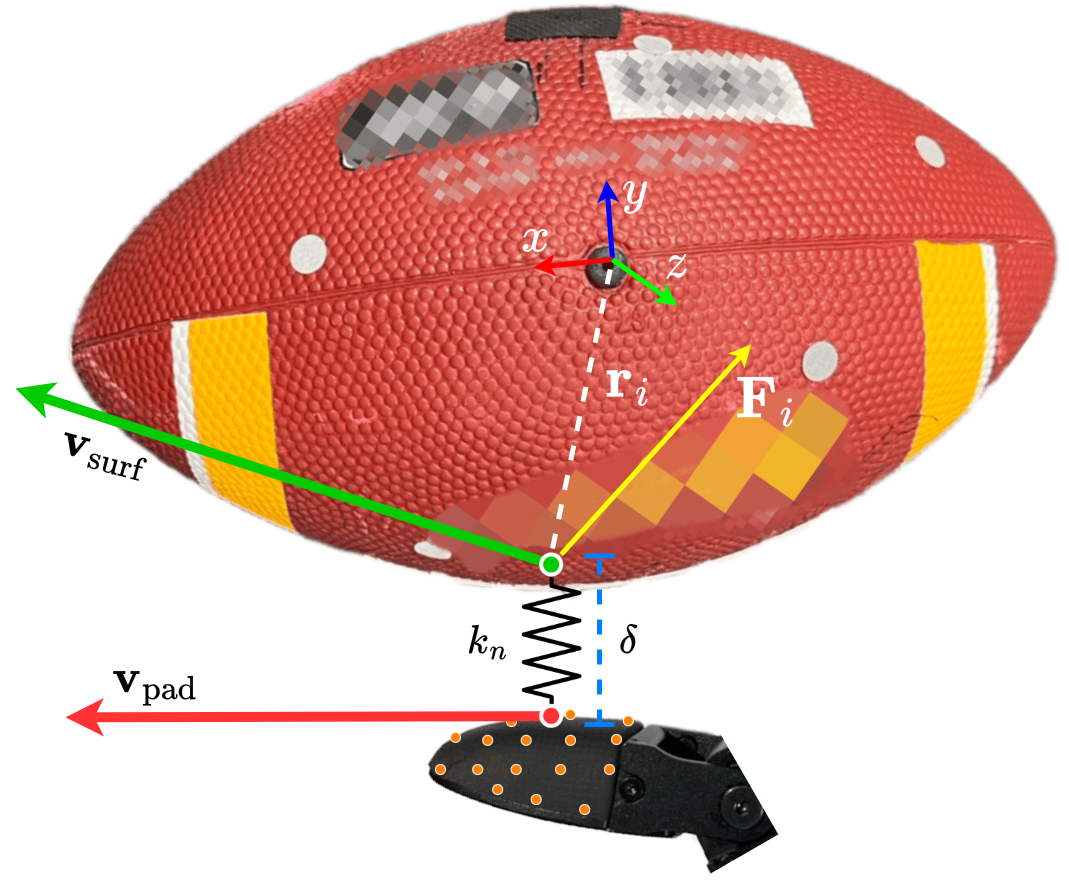}
    \vspace{-5mm}
    \caption{
    Illustration of the contact-aware football dynamics model, where $\vv_{\text{surf}}$, $\vv_{\text{pad}}$, and orange dots denote the contact point velocities of the football surface, the fingertip pad, and the sampled contact points, respectively.
    }
    \vspace{-8mm}
    \label{fig:football_dynamics}
\end{figure}

The contact-aware football dynamics $f_{\text{b}}$ in~\eqref{eq:follow_through_mpc} model how residual fingertip contacts affect the football after thumb opening. Although tactile sensors are integrated into the gripper, they are not used for force estimation because the transient follow-through contact lasts $\le 100\text{ ms}$ (occasionally a few milliseconds), resulting in unreliable sensor readings. Instead, contact forces are estimated using a differentiable geometric model that represents the football as a rigid superellipsoid. Each fingertip pad is discretized into surface samples, whose signed distances from the football are smoothly aggregated to estimate the effective contact point and penetration depth. The penetration depth is then mapped to contact force using a penalty-based spring model (illustrated in Fig.~\ref{fig:football_dynamics}). Specifically, for each sampled point on the pad, we evaluate the signed distance $\phi_n$ to the football surface, where $\phi_n<0$ indicates penetration. A softmax weighting over these sampled points yields a differentiable effective contact point and signed distance $\phi_{\text{soft}}$. The corresponding penetration depth is defined as $\delta=\max(0,-\phi_{\text{soft}})$ and converted into a normal contact force using the penalty-based spring model $\lambda=\operatorname{clip}(k_n\delta,0,\lambda_{\max})$, where $\lambda_{\max}$ represents the maximum allowable normal force, and $k_n$ is a tuned spring constant. With surface normal $\hat{\n}_{\text{s}}$, tangential relative velocity $\vv_{\text{t}}$, friction coefficient $\mu$, and a small regularizer $\varepsilon$, the fingertip force is $\mathbf{F}_i=-\lambda\hat{\n}_{\text{s}} -\mu\lambda\vv_{\text{t}}/(\|\vv_{\text{t}}\|+\varepsilon)$. The fingertip torque is $\tauu_i=\mathbf{r}_i\times\mathbf{F}_i$, where $\mathbf{r}_i$ is the moment arm from the football's center of mass to the effective contact point. The net external forces and torques acting on the football are $\fext=\sum_i\mathbf{F}_i$ and $\tauext=\sum_i\tauu_i$, respectively, and are integrated using the football rigid-body dynamics in~\eqref{eq:football_dynamics}. Through this contact-aware dynamics, the joint-velocity commands $\uu_{\text{rel},k} = \dq_{\text{rel}, k}$ govern the fingertip trajectories via forward kinematics, systematically altering the contact forces and predicting the football's next state.

\section{Experiment}\label{sec:exp}

In this section, we demonstrate the proposed methodology on a Unitree G1 humanoid robot equipped with a Unitree Dex3-1 gripper in the real world. We used the GR00T Whole-Body controller~\cite{bjorck2025gr00t} for control of the legged joints, while the upper body joints (including the 3-DoF waist) were controlled by the manipulation controller $\pi_{\text{mnp}}$ (Section~\ref{sec:method}). Before proceeding to the detailed hardware and sensing setup and experimental results, we first emphasize the primary observations as follows:

\begin{itemize}
    \item The throw phase, coupled with the legged controller and EKF-based pelvis-state estimation, exhibits consistent performance, establishing a reliable baseline for evaluating the subsequent follow-through phase.
    \item The follow-through phase actively improves the spiral quality, whereas uncontrolled transient finger-football interactions degrade it, as detailed in Section~\ref{subsec:follow-through-results}.
    \item Tight spiral throwing is heavily dominated by the shoulder and waist joints, providing a quantitative robotic analog to core-driven human throwing mechanics.
\end{itemize}

We first describe the hardware and sensing setup used to execute the humanoid football throw and quantify the resulting spiral (Section~\ref{subsec:hardware-sensing}), followed by evaluations of the throw phase (Section~\ref{subsec:throw-results}) and follow-through phase (Section~\ref{subsec:follow-through-results}).

\subsection{Hardware and Sensing Setup}
\label{subsec:hardware-sensing}
All experiments were conducted on a Unitree G1 29-DoF humanoid robot equipped with a Unitree Dex3-1 7-DoF dexterous gripper as shown in Fig.~\ref{fig:whole_body_control_architecture}. To enable a clean and controlled release of the football, the thumb was modified with a custom 3D-printed extended nail. 

We adopted a $0.252$~kg football inflated to an internal pressure of $12$~psi. It had a long-axis length of $21.6$~cm, a maximum diameter of $12.7$~cm, and a superellipsoid exponent of $n=1.66$.

The motions of the gripper and football were tracked in real time using reflective markers and an OptiTrack motion-capture (MoCap) system equipped with six high-speed Prime$^\text{X}$ 22 cameras. The system provided full 3D coverage of the throwing workspace, streamed data at $360$~Hz, and achieved sub-millimeter level tracking accuracy ($\leq 0.15$~mm). The raw gripper and football poses were independently filtered using body-specific Kalman filters.

\begin{table}[b]
\vspace{-6mm}
\centering
\caption{Controller Parameters}
\vspace{-3mm}
\label{tab:parameters}
\footnotesize
\renewcommand{\arraystretch}{0.8}
\setlength{\aboverulesep}{1pt}
\setlength{\belowrulesep}{1pt}

\begin{tabular}{@{} l @{\hspace{15pt}} l @{\hspace{40pt}} l @{\hspace{15pt}} l @{}}
\toprule
\textbf{Sym.} & \textbf{Value} & \textbf{Sym.} & \textbf{Value} \\ 
\midrule
\multicolumn{4}{@{}l}{\textit{Throw Phase}} \\ 
\midrule
$\mathbf{w}_{\text{run}}$ & $[10^3, 0.01, 0.05, 10^3]$ & $T$ & $0.60$~s \\
$\mathbf{w}_{\text{term}}$ & $[10^4, 2{\times}10^4, 2{\times}10^4]$ & $t_a$ & $0.55$~s \\
$\vv^*$ & $[3, 0, 0]$~m/s & $\w^*$ & $[-6, 0, 0]$~rad/s \\
$\text{w}_{\text{align}}$ & $10^6$ & &\\
\midrule
\multicolumn{4}{@{}l}{\textit{Follow-Through Phase}} \\ 
\midrule
$\mathbf{w}_{\text{mpc}}$ & $[10^3, 0.5, 0.02, 5.0]$ & $H$ & $15$ \\
$\Delta t$ & $0.004$~s & $v_{\text{safe}}$ & $0.05$~m/s \\
\midrule
\multicolumn{4}{@{}l}{\textit{Contact Aware Football Dynamics Model}} \\ 
\midrule
$k_n$ & $1800$ & $\lambda_{\max}$ & $12$~N \\
$\mu$ & $0.6$ & $\varepsilon$ & $10^{-6}$ \\ 
\bottomrule
\end{tabular}
\end{table}

\begin{figure*}[t]
    \centering
    \vspace{2mm}
    \includegraphics[width=\textwidth]{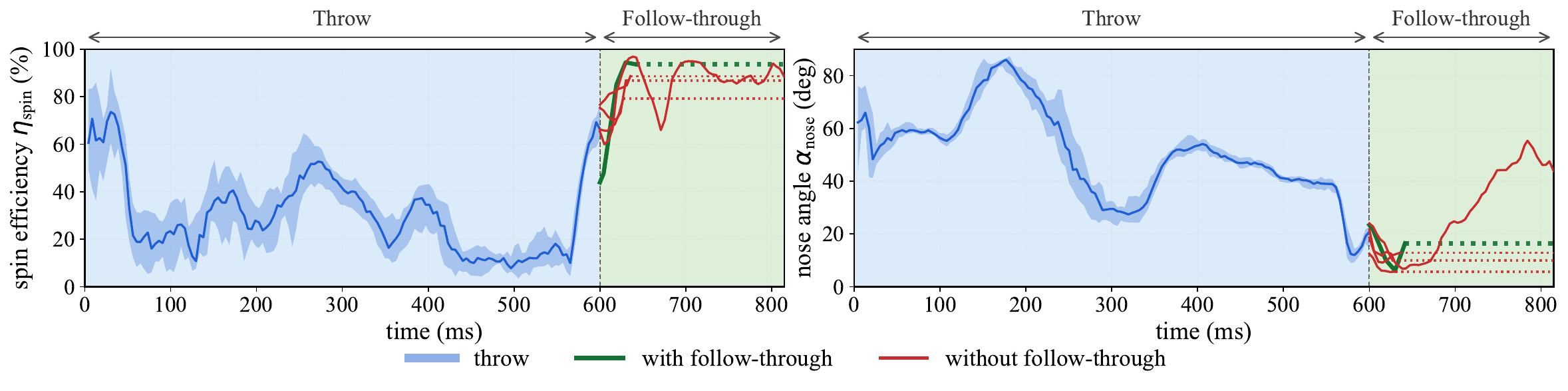}
    \vspace{-10mm}
    \caption{
    Temporal evolution of spin efficiency $\spin$ (left) and nose-velocity angle $\alphan$ (right) throughout multiple trials. The blue shaded band represents the distribution of multiple attempts of the throw phase using the presented method in Section~\ref{sec:throw_phase}. At the point of release, follow-through ensures a clean finger separation, preserving high spin efficiency (shown by the green colored line). Conversely, without follow-through, uncontrolled release induces a pronounced wobble, characterized by decaying spin efficiency and progressive misalignment (shown in red). Each trace terminates upon complete finger separation and is held constant thereafter for visual comparison.
    }
    \vspace{-8mm}
    \label{fig:spiral_comparison}
\end{figure*}

\subsection{Throw Phase Results}
\label{subsec:throw-results}
The continuous-time trajectory optimization problem~\eqref{eq:trajopt} was solved using direct multiple shooting. We discretized the trajectory into 50 steps (over the 0.6~s horizon) and optimized it offline using Crocoddyl, with Pinocchio providing the robot dynamics. The trajectory optimization converged in $40$ iterations and $0.11$~s, with no observed constraint violations. The optimization was performed on a reduced $10$-DoF model comprising the waist and throwing-arm joints. The target orientation $R_{\text{target}}$ was computed by aiming the football straight forward at a $35^\circ$ launch elevation, with the remaining parameters listed in Table~\ref{tab:parameters}. The trajectory optimization was solved using Crocoddyl's box-constrained, feasibility-driven differential dynamic programming solver. The resulting throw phase demonstrated consistent performance across all real-world trials. At the terminal boundary of the throw phase, the football had a mean spin efficiency $\spin$ of $67.1\% \pm 12.1\%$, and a nose-angle error $\alphan$ of $18.9^\circ \pm 5.6^\circ$ (across 17 trials). This degradation from the planned trajectory is due to actuator overheating in each trial, gradual hardware wear over time, and highly non-linear joint dynamics at high velocities. For example, during the final 50~ms of one throw, the waist pitch joint was saturated for $90.9\%$ of the interval, while the shoulder pitch and elbow joints were each saturated for $54.5\%$.

Although the throw phase produces a consistent terminal state, the football is not yet fully released because the fingertips remain in contact with its surface. The subsequent follow-through phase therefore regulates these residual contacts until complete detachment to prevent contact-induced disturbances from degrading the spiral.

\subsection{Follow-Through Phase Results}
\label{subsec:follow-through-results}
Real-world hardware experiments highlighted the significance of the follow-through phase for the final release state. The MPC problem was solved via a sampling based solver. The controller and contact-dynamics parameters are listed in Table~\ref{tab:parameters}. A single online MPC solve required approximately $500$~ms. This computation time is unfeasible for the $250$~Hz control frequency required for such a highly dynamic task. Consequently, the online follow-through phase used a lookup-table MPC, where the underlying lookup table was constructed offline from real-world collected data. The table uses a $52$-dimensional state key encoding ball spin, contact geometry, relative motion, wrist-finger state, elbow motion, elbow pose, grip offset, and gravity. Across 61 trials, the follow-through phase lasted $78.5$~ms on average. In some trials, it lasted only $30$-$40$~ms. At the point of complete detachment, the football achieved a peak linear velocity of 5.35~m/s and an angular velocity of 14.5~rad/s. With follow-through, properly controlling the fingertip and football interaction during release yielded up to $93.6\%$ spin efficiency $\spin$ and a nose-angle error $\alphan$ of $16.4^\circ$. Without an active follow-through phase, uncontrolled transient finger–football interactions significantly disturbed the release, causing the final detachment state to deviate from the terminal state of the throwing phase. Consequently, average spin efficiency decreased from $67.1\%$ to $60.9\%$, while average nose-angle error increased from $18.9^\circ$ to $34.7^\circ$. In the worst-case observed trial, spin efficiency collapsed to $18.5\%$, and the nose-angle error increased to 48.5$^\circ$. Figure~\ref{fig:spiral_comparison} compares the spiral metrics for multiple throws with and without an active follow-through phase. These results highlight that treating the throw as an instantaneous release substantially degrades spiral quality, making a controlled follow-through phase essential to both preserve spiral quality and improve spin efficiency.

\begin{figure}[b]
    \centering
    \vspace{-6.5mm}
    \includegraphics[width=0.98\columnwidth]{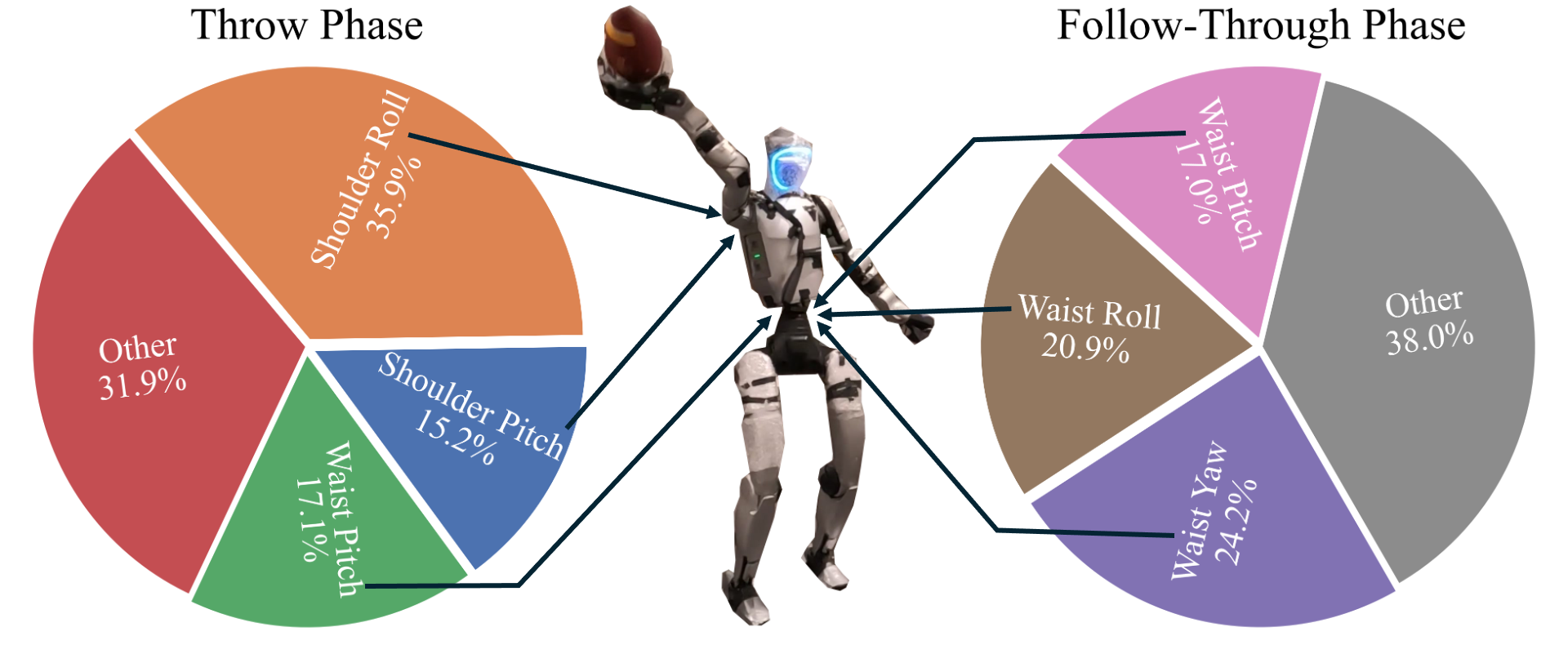}
    \vspace{-4mm}
    \caption{
    Joint-level torque impulse distribution for the throw and follow-through phases. Each pie chart shows the normalized time integral of joint torque, with the top three contributing joints highlighted. The throw phase is dominated by shoulder roll, waist pitch, and shoulder pitch, while the follow-through phase is dominated by waist yaw, waist roll, and waist pitch.
    }
    \label{fig:joint_torque_impulse}
\end{figure}

\subsection{Insights into Human Throwing Mechanics} 
\label{subsec:human-insights}
Beyond robotic control, the torque profiles generated by the proposed framework offer compelling insights that parallel human quarterback mechanics. The integrated commanded torques in Fig.~\ref{fig:joint_torque_impulse} indicate that the shoulder roll, shoulder pitch, and waist pitch joints contribute most strongly to momentum generation during the throw phase. Conversely, during the highly sensitive follow-through phase, the kinetic chain shifts its reliance toward the waist yaw, roll, and pitch joints. These emerge as the primary contributors to maintaining and regulating the prolonged fingertip contact necessary for active alignment correction during the follow-through phase. The prominent role of the waist is consistent with quarterback training principles that emphasize coordinated torso rotation and core stability for generating and controlling a tight spiral throw. However, an interesting biomechanical distinction arises from the humanoid's kinematic topology: while humans have highly coupled, non-independent degrees of freedom, the robot possesses independent, orthogonal waist joints. Consequently, the trajectory optimization systematically leverages these decoupled joints to preserve the spiral, demonstrating how an optimal control framework mathematically decomposes and solves a biomechanical task that human athletes must execute as a complex, coupled physical motion. Another notable distinction is the absence of the rapid wrist pronation typically seen in human throws. This may, in part, be related to the short time duration of the follow-through phase and, in part, to the absence of a spin-acceleration term in the MPC formulation. 

\section{Conclusion}
\label{sec:conclusion-limitations}
This paper presents, to the best of our knowledge, the first work to enable a humanoid robot to execute a tight spiral American football throw. Recognizing that the gripper release is a transient dynamic process rather than an instantaneous event, we introduce a sequential control strategy: a primary throw phase to generate the coupled momentum, followed by a dedicated follow-through phase to regulate finger-ball separation and stabilize the tight spiral. We validated this methodology on a Unitree G1 humanoid equipped with a Dex3-1 gripper, utilizing a motion capture system to guide the follow-through phase and rigorously quantify spiral metrics. Through an ablation study, we demonstrated the critical necessity of the follow-through phase. Compared to an uncontrolled release, the active follow-through not only preserved the spiral quality established during the throw, but substantially improved it leading up to complete detachment. Furthermore, our analysis showed that the waist and shoulder joints contribute most strongly to the momentum required for a tight spiral. Due to the challenges mentioned above (e.g., overheating and hardware degradation), the current evaluation does not yet support a statistically significant repeatability
analysis, which we will improve in future works.

Future work will develop a unified whole-body controller that leverages the robot's full-body inertia to generate higher momentum. This approach will distribute physical effort across all the joints, enabling higher release velocities while further improving spiral quality. Additionally, we will incorporate a spin-acceleration term into the MPC formulation and integrate an anthropomorphic, five-fingered hand to better replicate human dexterity.
\printbibliography

\end{document}